\documentclass{article}
\usepackage{spconf,amsmath,graphicx}

\usepackage{enumitem}
\setlist{nosep, leftmargin=14pt}
\usepackage{nohyperref}
\usepackage{mwe} % to get dummy images
\usepackage{adjustbox}
\usepackage{booktabs}
\usepackage[table,x11names,dvipsnames,table]{xcolor}

\usepackage{graphicx}
\usepackage{dblfloatfix} % improves two-column float placement
\usepackage{booktabs}

\title{Catena: A Comprehensive Software Suite for Large-Scale Connectomics}
\newcommand*{\affaddr}[1]{#1} % No op here. Customize it for different styles.
\newcommand*{\affmark}[1][*]{\textsuperscript{#1}}

\usepackage{amssymb}% http://ctan.org/pkg/amssymb
\usepackage{pifont}% http://ctan.org/pkg/pifont

\usepackage{listings}
\usepackage{mathptmx} % Times font for compatibility

\newcommand\YAMLcolonstyle{\color{red}\mdseries}
\newcommand\YAMLkeystyle{\color{black}\bfseries}
\newcommand\YAMLvaluestyle{\color{blue}\mdseries}

\makeatletter

\newcommand\language@yaml{yaml}

\expandafter\expandafter\expandafter\lstdefinelanguage
\expandafter{\language@yaml}
{
  keywords={true,false,null,y,n},
  keywordstyle=\color{darkgray}\bfseries,
  basicstyle=\small,                                 % assuming a key comes first
  sensitive=false,
  comment=[l]{\#},
  morecomment=[s]{/*}{*/},
  commentstyle=\color{purple}\ttfamily,
  stringstyle=\YAMLvaluestyle\ttfamily,
  moredelim=[l][\color{orange}]{\&},
  moredelim=[l][\color{magenta}]{*},
  moredelim=**[il][\YAMLcolonstyle{:}\YAMLvaluestyle]{:},   % switch to value style at :
  morestring=[b]',
  morestring=[b]",
  literate =    {---}{{\ProcessThreeDashes}}3
                {>}{{\textcolor{red}\textgreater}}1     
                {|}{{\textcolor{red}\textbar}}1 
                {\ -\ }{{\mdseries\ -\ }}3,
}

\lst@AddToHook{EveryLine}{\ifx\lst@language\language@yaml\YAMLkeystyle\fi}
\makeatother

\newcommand\ProcessThreeDashes{\llap{\color{cyan}\mdseries-{-}-}}

\name{%
\begin{tabular}{@{}c@{}}
Samia Mohinta\thanks{$\dagger$ Corresponding author: sm2667@cam.ac.uk}\affmark[1,2,4, $\dagger$],
Pedro Gómez-Gálvez\affmark[2,3],
Shi Yan Lee\affmark[1,2],
Daniel Franco-Barranco\affmark[2,5], \\
Michael Clayton\affmark[2],
Stephan Preibisch\affmark[4],
Jan Funke\affmark[4],
Albert Cardona\affmark[2,1]
\end{tabular}}

\address{
\affaddr{\affmark[1] Dept. of Physiology, Development and Neuroscience, University of Cambridge, UK} \\
\affaddr{\affmark[2] MRC Laboratory of Molecular Biology, UK} \\
\affaddr{\affmark[3] Instituto de Biomedicina de Sevilla (IBiS), Hospital Universitario Virgen del Rocío/CSIC/\\Universidad de Sevilla and Dept. de Biología Celular, Facultad de Biología, Universidad de Sevilla, Spain} \\
\affaddr{\affmark[4] HHMI Janelia, USA} \\
\affaddr{\affmark[5] Donostia International Physics Center (DIPC), San Sebastian, Spain}
}
\begin{document}
%\ninept
%
\maketitle
\begin{abstract} %164 words
The gold standard datasets for mapping connectomes are electron microscopy volumes of densely labeled neural tissue at nanometer resolution. Yet reconstructing and proofreading neuronal arbors and annotating all synapses requires pipelining multiple software tools that are often fragmented, inconsistently maintained, or proprietary, hindering reproducibility and automation. Here, we introduce Catena, an open-source, comprehensive, developer-centric software suite for connectomics that integrates modules for 3D neuron and organelle segmentation, synapse detection, microtubule tracking, and neurotransmitter inference. Catena organizes its modules in composable, chunk-wise processing pipelines in a completely documented, extensible, and adaptable design. We further reduce compute and ground-truth data requirements with pretrained machine learning models, facilitating fine-tuning. Catena ships fully containerized modules that encapsulate evolving dependencies for consistent execution across workstations and clusters. By consolidating open components, shareable models, and containerized runtimes, Catena delivers a reproducible and scalable approach to mapping cellular connectomes from electron microscopy volumes. Code and documentation: https://github.com/Mohinta2892/catena.git

\end{abstract}


\section{INTRODUCTION}
\label{sec:intro}

%Connectomics seeks to reconstruct wiring diagrams (connectomes) at synaptic resolution to link structure to function (behavior) in both healthy and pathological brains.

Recent advances in volume electron microscopy (EM) have made brain-scale imaging practical, leading to mapping complete, dense connectomes in invertebrates like \emph{Drosophila}~\cite{Winding2023TheBrain_edit, Dorkenwald2024NeuronalBrain_edit} 
% Takemura2024_edit, 
and \emph{C. elegans}~\cite{witvliet2021connectomes} and for partial volumes in vertebrates, such as in the mouse visual cortex~\cite{Bae2021FunctionalCortex_edit}) and human neocortex~\cite{Shapson-Coe2024AResolution_edit}. 
% Historically, landmark reconstructions were achieved largely by manual annotation and proofreading~\cite{Winding2023TheBrain_edit, White1986TheElegans, Eichler2017TheBrain_edit} but such workflows are prohibitively slow at scale; for perspective, it took almost 10 years to reconstruct a larval fly brain a mouse brain is roughly three orders of magnitude larger than a fruit fly brain, compounding both data volume and proofreading burden.
Historically, landmark connectomes were mapped largely through manual annotation and proofreading~\cite{White1986TheElegans, Ohyama2015ADrosophila_edit}, a workflow that becomes prohibitively slow at scale. For perspective, the larval \textit{Drosophila} connectome ($\sim$3016 neurons) required nearly a decade of manual effort~\cite{Winding2023TheBrain_edit}; by comparison, a mouse brain contains $\sim$70 million of much larger neurons, dramatically increasing both data volume and proofreading burden\footnote{https://wellcome.org/insights/reports/scaling-connectomics}.

While deep learning has produced powerful 3D segmentation models for biomedical datasets including EM~\cite{Ronneberger2015U-Net:Segmentation,Lee2017SuperhumanChallenge,  Januszewski2018High-precisionNetworks_edit, Sheridan2023LocalSegmentation_edit}, most solutions remain task-specific, difficult to reproduce, or hard to extend. Community efforts such as BiaPy~\cite{Franco-Barranco2025BiaPy:Bioimages_edit}, ZeroCostDL4Mic~\cite{vonChamier2021DemocratisingZeroCostDL4Mic_edit} and Cellpose~\cite{Stringer2021Cellpose:Segmentation} have lowered entry barriers, but few frameworks are engineered for robust, end-to-end processing of multi-terabyte EM volumes. DaCapo~\cite{Patton2024DaCapo:Segmentation_edit}, for example, is a notable step forward but focuses primarily on segmentation.

Here, we introduce Catena, a comprehensive, developer-centric software suite designed to address these challenges. In contrast to disparate tools, our \emph{primary contribution} is the integration of state-of-the-art modules for core connectome mapping (neuron segmentation, synapse detection) and biological enrichment (mitochondria, microtubules, neurotransmitters) into a single, scalable pipeline. Our approach is built on three key principles: \textbf{(1) Scalability}, using \emph{composable, chunk-wise} workflows for multi-terabyte volumes; \textbf{(2) Reproducibility}, via fully containerized runtimes that insulate users from dependency drift; and \textbf{(3) Accessibility}, through the release of open-weights pretrained models and datasets. Together, these features ensure reproducible, large-scale connectome mapping across heterogeneous computing environments and imaging protocols.

\begin{figure*}[!ht]
\centering
 \includegraphics[width=1\textwidth]{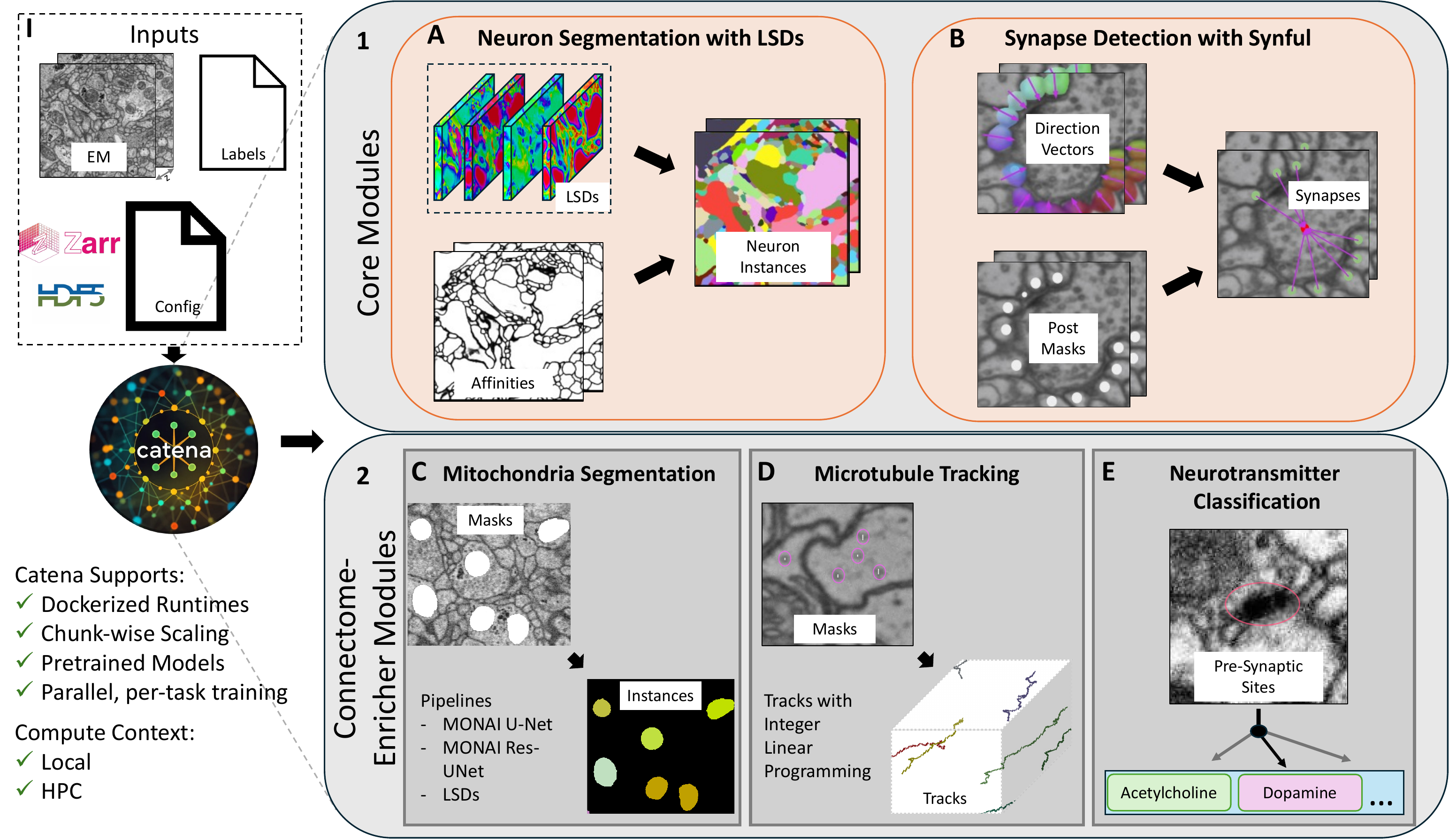}
  \vspace{-0.7cm}
 \caption{\textbf{Schematic of Catena for scalable automated connectome mapping}. From \textbf{(I)} raw EM inputs, Catena orchestrates two main stages: \textbf{(1)} \textbf{Core Modules} for segmenting neuron instances and detecting synaptic partners, and \textbf{(2)} \textbf{Connectome-Enricher Modules} for further annotations, including mitochondria segmentation, microtubule tracking and neurotransmitter classification. Catena offers modular, Dockerized runtimes, chunk-wise scaling, pretrained models, and parallel task execution across both local and HPC compute environments.}
 \label{fig:catena_schematic}
\end{figure*}

\section{METHOD}
\label{sec:method}
Catena addresses scalability, reproducibility, and usability in connectomics via a fully open-source, open-weights design. We release pretrained models in PyTorch and TensorFlow together with accompanying datasets to support reuse and end-to-end reproducibility. The software suite's modular architecture supports workflows from raw EM input to an enriched connectome (Fig.~\ref{fig:catena_schematic}), either by running the full pipeline or by composing only any modules as needed.

\subsection{Modular Architecture: Catena Modules}
Each module exposes clear inputs/outputs and participates in chunk-wise workflows suitable for large volumes. The modules fall into two groups: core connectome extraction and connectome enrichment. Catena also integrates proofreading tools to manually verify and correct model predictions.

\subsubsection*{Core extraction modules.}
\begin{itemize}
  \item \textbf{Neuron segmentation, Fig.~\ref{fig:catena_schematic}A.} We implement Sheridan et al.'s Local Shape Descriptors (\texttt{LSDs})~\cite{Sheridan2023LocalSegmentation_edit} for volumetric EM, jointly predicting voxel-wise 3D affinities and per-voxel shape descriptors (e.g., neuron shape, elongation, orientation). This coupling improves long-range trajectory continuity across large volumes and brings affinity-based segmentation to parity with state-of-the-art flood-filling networks~\cite{Januszewski2018High-precisionNetworks_edit}, while delivering roughly 100× speed up. The predicted affinities are converted to instances via a watershed–agglomeration pipeline using Waterz\footnote{{https://github.com/funkey/waterz}}, wherein (i) affinities are thresholded to obtain a foreground mask; (ii) a distance transform is computed and local maxima are detected; (iii) these maxima seed a watershed to produce an over-segmentation (supervoxels); (iv) a region adjacency graph is built by connecting touching supervoxels; and (v) edges are hierarchically agglomerated in order of decreasing affinity until a validation-tuned stopping threshold is reached.
  \item \textbf{Synapse detection, Fig.~\ref{fig:catena_schematic}B.}
  We implement Buhmann et al.'s \texttt{Synful}~\cite{Buhmann2021AutomaticSet_edit}, a single-stage 3D U-Net for joint synapse detection and partner identification. In contrast to traditional two-step methods, \texttt{Synful} simultaneously predicts a voxel-wise postsynaptic mask and regresses a 3D direction vector from each postsynaptic voxel to its corresponding presynaptic site. \texttt{Synful} can be trained using only sparse point annotations, independent of a full neuron segmentation, which significantly reduces the manual annotation burden. Predicted connections are then extracted via a post-processing pipeline wherein: (i) the postsynaptic mask is thresholded to find connected components; (ii) components are filtered based on a connection score; (iii) a postsynaptic location is identified within each valid component; and (iv) the corresponding presynaptic location is recovered by following the regressed direction vector. Finally, we attach these predicted synaptic partners to their corresponding neuron segments.
  
  % We implement Buhmann et al.'s \texttt{Synful}~\cite{Buhmann2021AutomaticSet_edit}, a single-stage 3D U-Net that predicts post-synaptic masks and regresses post-to-pre synaptic direction vectors from point annotations of synapses in EM volumes, recovering post- and pre-synaptic sites simultaneously. Synful can be trained with and without the availability of neuron segmentation, thereby giving flexibility to end users and reducing the annotation burden. We further apply post-processing to attach predicted synapses to their corresponding neuron segments.
\end{itemize}

\noindent
To broaden adoption and simplify maintenance, we re-implement the TensorFlow 1.x versions of \texttt{LSDs} and \texttt{Synful} in PyTorch, extending their 3D U-Net architectures with optional batch normalization to support stable training with larger batch sizes.

\subsubsection*{Connectome-enrichment modules.}
% \vspace{-0.1cm}
Enrichment modules augment the core wiring diagram with biological context, providing crucial evidence to inform proofreading of neuron paths and connectivity, and enabling deeper structure-function analysis.

\begin{itemize}
  \item \textbf{Organelle (mitochondria) segmentation, Fig.~\ref{fig:catena_schematic}C.} For mitochondria segmentation, we employ MONAI\footnote{https://github.com/Project-MONAI/MONAI}-based 3D U-Net~\cite{Cicek20163DAnnotation} pipelines, leveraging its robust, domain-specific tools for a reproducible baseline, and adapt Xie et al.'s~\cite{Xie2024TransferTissues_edit} Residual U-Net, chosen for its proven efficiency with transfer learning on EM data. Furthermore, we extend the \texttt{LSDs} to enable a novel, joint, multi-class segmentation of both neurons and mitochondria when ground-truth labels are available for both. This module is critical as mitochondrial morphology is intrinsically linked to cellular function and disease, including synaptic efficacy~\cite{Perkins2010}.
  % For mitochondria segmentation, we employ MONAI\footnote{{https://github.com/Project-MONAI/MONAI}}-based 3D U-Net pipelines and adapt Xie et al.'s~\cite{Xie2024TransferTissues_edit} Residual U-Net for efficient training on electron micrographs. Furthermore, we extend the \texttt{LSDs} to enable joint, multi-class segmentation of both neurons and mitochondria when ground-truth labels are available for both. This module is critical as mitochondrial morphology is intrinsically linked to cellular function and disease.
  \item \textbf{Microtubule tracking, Fig.~\ref{fig:catena_schematic}D.} 
  To enrich the connectome with cytoskeletal structure which can aid in proofreading neurites, Catena implements the hybrid microtubule tracking method from Eckstein et al.~\cite{Eckstein2020MicrotubuleVolumes_edit}. A 3D U-Net first localizes microtubule candidates voxel-wise, after which a constrained Integer Linear Program (ILP) recovers continuous trajectories by applying biological priors like limited branching. This method’s highly efficient ILP formulation provides orders-of-magnitude speed improvements over prior methods and supports the block-wise processing required for terabyte-scale volumes.  
  % Microtubule tracking provides additional structural information as they are part of the cytoskeleton of cells and can therefore aid in proofreading and continuity checks along neurites. Within Catena, we re-implement Eckstein et al.~\cite{Eckstein2020MicrotubuleVolumes_edit}'s 3D U-Net to first localize microtubules voxel-wise; next convert these segmentations into an undirected graph connecting nearby candidates. By encoding local and biological constraints (limited branching and curvature), trajectories are recovered via solving a constrained optimization problem with an Integer Linear Program.
  \item \textbf{Neurotransmitter classification, Fig.~\ref{fig:catena_schematic}E.} We adapt Synister~\cite{Eckstein2024NeurotransmitterMelanogaster_edit}, a pioneering work that demonstrated how artificial feedforward networks~\cite{Simonyan2015VeryRecognition} can reliably classify transmitters from subtle ultrastructural EM cues. We use presynaptic locations with categorical neurotransmitter labels (e.g., acetylcholine, dopamine, glutamate) to classify the neurotransmitter signature of synaptic connections, thereby inferring their putative functional sign (e.g., excitatory or inhibitory) at the neuron level.
  
  % We adapt Synister~\cite{Eckstein2024NeurotransmitterMelanogaster_edit} to train a 3D VGG model and include a ResNet baseline. We use presynaptic locations with categorical neurotransmitter labels (e.g., acetylcholine, dopamine, glutamate) to classify the transmitter identity of connections, thereby inferring their putative functional sign (e.g., excitatory or inhibitory) at the neuron level.
  % We adapt Synister~\cite{Eckstein2024NeurotransmitterMelanogaster_edit} for supervised training of a 3D VGG model and include a ResNet baseline. These models use pre-synaptic locations and categorical labels (e.g., acetylcholine, dopamine, glutamate) as inputs to assign putative functional classes (e.g., excitatory/inhibitory) to neuron-to-neuron connections indicating information flow in biological brains.
\end{itemize}

  \begin{figure}[ht!]
\centering
 \includegraphics[width=0.48\textwidth]{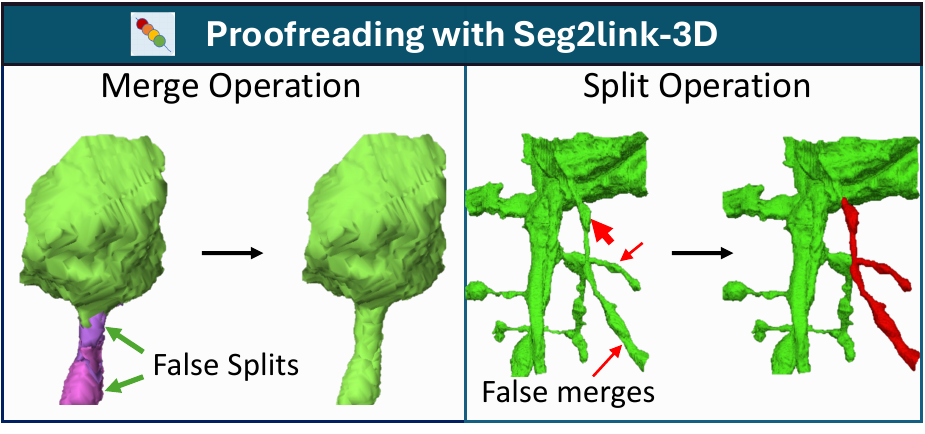}
 \vspace{-0.7cm}
 \caption{\textbf{Proofreading neuron segmentations with Seg2link-3D.} False splits are corrected with \textbf{merge} operations (left), and false merges with \textbf{split} operations (right).}
 \label{fig:seg2link}
\end{figure}
% \vspace{-0.4cm}

\subsubsection*{Proofreading.}
Catena integrates Seg2Link~\cite{Wen2023Seg2Link:Stacks}, a semi-supervised, \texttt{napari}-based tool~\cite{Sofroniew2025-xr}, and extends it with native support for Zarr datasets for manual correction of merge/split errors in 3D neuron segmentations in small volumes (Fig.~\ref{fig:seg2link}). For large-scale review, Catena provides CAVE~\cite{Dorkenwald2025CAVE:Engine_edit} and CATMAID~\cite{Saalfeld2009CATMAID:Data} scripts that enable whole-brain inspection of meshes and skeletons, respectively. For synapses, Catena includes import scripts to load \texttt{Synful} predictions into CATMAID, enabling systematic, dataset-scale proofreading of synaptic connectivity.

\subsection{Catena Capabilities and Assets}
\textbf{Pretrained Models, Datasets, and Training Runtimes.}  Catena provides pretrained weights with corresponding datasets for fine-tuning or direct use. Accepted data formats include Zarr and HDF5 (Fig.~\ref{fig:catena_schematic}I), which store multi-dimensional data in chunks for efficient I/O. Documentation specifies system requirements, data organization for training and inference, expected data types and formats, and output conventions, and includes guidance for adapting modules to new datasets (e.g., architecture changes, patch sizes) plus troubleshooting notes collected across multiple compute infrastructures. Catena handles both isotropic and anisotropic EM volumes.

Catena uses the Gunpowder library\footnote{{https://github.com/funkelab/gunpowder}} for its core data processing tasks, including data loading and augmentation, as its ability to operate in world units (nanometers) is crucial for correctly handling the anisotropy common in EM volumes. We also provide an optional integration with Weights \& Biases\footnote{{https://wandb.ai/site/}} for experiment tracking and logging.

% For data pipelines and training, Catena uses Gunpowder\footnote{{https://github.com/funkelab/gunpowder}} for data loading, augmentation, model checkpointing, and intermediate artifact management with Weights \& Biases\footnote{{https://wandb.ai/site/}} integration for online/offline monitoring. Gunpowder provides EM-aware transforms and supports voxel-resolution-aware augmentation at nanometer scale. Where voxel-space augmentation pipelines are preferable for small curated subsets, Catena also integrates MONAI-based workflows for ad hoc mitochondria segmentation and neurotransmitter classification.

% \textbf{Data augmentation via EM--to--EM style transfer.} To reduce appearance differences across datasets during training or inference, we implement a 2D CycleGAN~\cite{Zhu2020UnpairedNetworks} adapted to EM. Translated slices undergo quantitative and manual quality control before being re-stacked into 3D cubes for training Catena modules (Fig.~\ref{fig:em2em}).

% \begin{figure}[ht!]
% \centering
%  \includegraphics[width=0.48\textwidth]{fig_catena/config.png}
%  \vspace{-0.7cm}
%  \caption{\textbf{Example YACS-based \texttt{config.py} for Catena.} Configs control system, data, training, and model settings (including isotropic/anisotropic options), and enable reproducible, portable runs.}
%  \label{fig:config_example}
% \end{figure}

\vspace{0.2cm}
\noindent
\textbf{Evaluation and Analysis.}
Each Catena module ships with task-specific evaluation scripts that operate on chunked predictions and ground truth, aggregating scores across chunks to yield dataset-level metrics. For segmentation, we report Variation of Information (VOI) and Adapted Rand Index Error (ARAND); for detection/classification and tracking tasks, we report F1, precision, and recall. See Table~\ref{tab:metrics_overview} for the metrics used per module and their optimization direction ($\uparrow$/$\downarrow$).

% \documentclass{article}
% \usepackage{booktabs} % Required for \toprule, \midrule, \bottomrule

% \begin{document}

% \begin{table}[t]
% \rowcolors{2}{gray!10}{white} % odd rows light gray, even white
%   \caption{Primary evaluation metrics per module of Catena.}
%   \label{tab:metrics_overview}
%   \centering
%   \scriptsize
%   \setlength{\tabcolsep}{3pt}%
%   \renewcommand{\arraystretch}{1.05}
%   \begin{tabular}{@{} l >{\raggedright\arraybackslash}p{0.60\columnwidth} @{}}
%     \toprule
%     \textbf{Task} & \textbf{Metrics} \\
%     \midrule
%     Neuron Segmentation            & Variation of Information (VOI)$\downarrow$, Adapted Rand Index Error (ARAND)$\downarrow$ \\
%     Synapse Detection              & F1$\uparrow$, Precision$\uparrow$, Recall$\uparrow$ \\
%     Microtubule Tracking           & F1$\uparrow$ (node \& topology), Precision$\uparrow$, Recall$\uparrow$ \\
%     Neurotransmitter Classification& F1$\uparrow$ (per class), Precision$\uparrow$, Recall$\uparrow$ \\
%     Mitochondria Segmentation      & \emph{Semantic:} IoU$\uparrow$, Dice$\uparrow$; \emph{Instance:} mAP$\uparrow$, F1$\uparrow$ \\
%     \bottomrule
%   \end{tabular}
%   \vspace{2pt}
%   {\footnotesize ↑ higher is better; ↓ lower is better.}
% \end{table}

\begin{table}[t]
\centering
\scriptsize
\caption{Primary evaluation metrics per module of Catena.}
\label{tab:metrics_overview}
\vspace{0.2cm}
\rowcolors{3}{gray!10}{white} % odd rows light gray, even white
\begin{tabular}{p{0.25\linewidth}p{0.65\linewidth}}
\toprule
Task & Metrics \\
\midrule
Neuron segmentation &
Variation of Information (VOI)$\downarrow$, Adapted Rand Index Error (ARAND)$\downarrow$, Expected Run Length (ERL) $\uparrow$\\

Synapse detection &
F1$\uparrow$, Precision$\uparrow$, Recall$\uparrow$ \\

Microtubule tracking &
F1$\uparrow$ (node and topology), Precision$\uparrow$, Recall$\uparrow$ \\

Neurotransmitter cls. &
F1$\uparrow$ (per class), Precision$\uparrow$, Recall$\uparrow$ \\

\begin{tabular}[l]{@{}l@{}}Mitochondria\\ segmentation\end{tabular} &
\begin{tabular}[l]{@{}l@{}}\underline{Semantic:} IoU$\uparrow$, Dice$\uparrow$ \\ \underline{Instance matching:} TP$\uparrow$/FP$\downarrow$/FN$\downarrow$ at IoU $\geq 0.5$\\ \quad association errors: one-to-one$\uparrow$/over-seg.$\downarrow$\\ \quad under-seg.$\downarrow$/many-to-many$\downarrow$/miss$\downarrow$ 

% background$\downarrow$

\end{tabular} \\
\bottomrule
\end{tabular}

\vspace{0.25em}
\footnotesize{$\uparrow$ higher is better; $\downarrow$ lower is better.}
\end{table}

% \end{document}

\

\begin{figure}
\centering
 \includegraphics[width=0.48\textwidth]{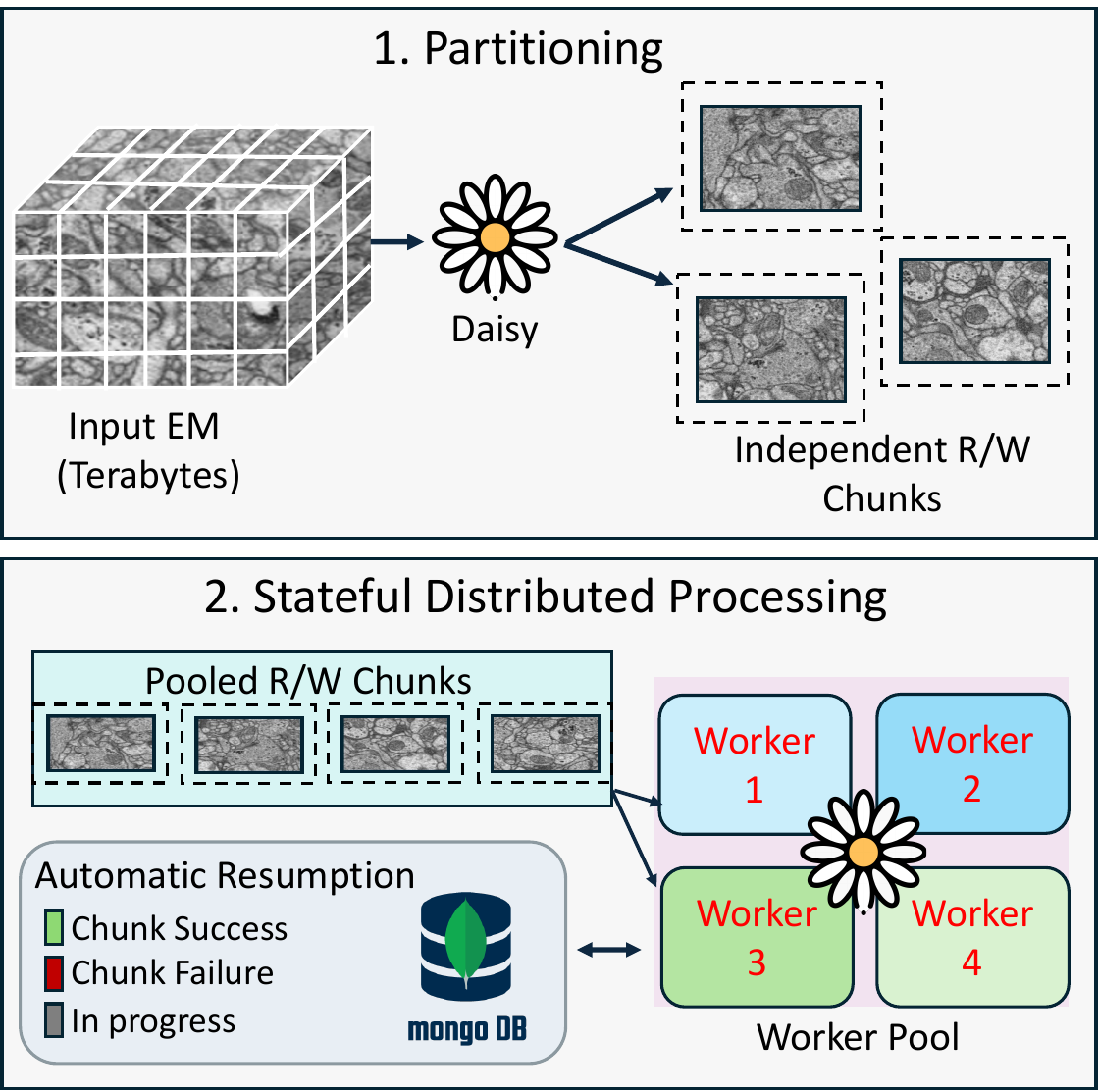}
 \vspace{-0.7cm}
 \caption{\textbf{Scalable, stateful chunk-wise processing in Catena.} \textbf{(1)} A terabyte-scale input EM volume is partitioned by Daisy into independent read/write chunks. \textbf{(2)} Pooled chunks are dispatched across a worker pool for distributed execution, with per-chunk status (success, failure, in progress) tracked to enable automatic resumption of failed tasks.}
 \label{fig:daisy_schematic}
\end{figure}

\noindent
\textbf{Scalable, Resumable, Chunk-wise Processing.}
% Within Catena, we deploy Daisy\footnote{{https://github.com/funkelab/daisy}} as the task scheduler and deterministic chunking library. Daisy partitions each volume into independent spatial chunks with fixed read/write windows, which are then dispatched to multiple workers, either on a single multi-GPU node or across an HPC cluster via SLURM, while a persistent MongoDB backend records per-chunk, per-stage status (succeeded/failed). Fig.~\ref{fig:daisy_schematic} illustrates this stateful, distributed workflow. Together, these components enable exact resumption from the last recorded state and targeted retries without reprocessing completed work, while keeping memory and I/O bounded by operating on windows rather than entire volumes.
Within Catena, we deploy Daisy\footnote{{https://github.com/funkelab/daisy}} as the task scheduler and deterministic chunking library \emph{during inference} on terabyte-scale EM volumes. Daisy partitions each volume into independent spatial chunks with fixed read/write windows and orchestrates their dispatch to workers (single multi-GPU node or HPC via SLURM). For persistent state tracking (per-chunk, per-stage status), Catena uses a backend database to record success/failure/progress. Fig.~\ref{fig:daisy_schematic} illustrates this stateful distributed workflow. Together, these components enable exact resumption from the last recorded state and targeted retries without reprocessing completed work, while keeping memory and I/O bounded by operating on chunks rather than entire volumes.

\

\noindent
\textbf{Reproducibility and Portability.} To insulate execution from dependency drift on workstations and clusters, Catena ships fully containerized runtimes\footnote{{https://hub.docker.com/u/mohinta2892}} per module. Each module is dockerized independently to prevent dependency overwrites or leakage during dependency updates and maintenance. Configuration 
% (Fig.~\ref{fig:config_example}) 
is handled via YACS\footnote{{https://github.com/rbgirshick/yacs}}, versioned as per the used training dataset.  All training modules use deterministic seeds for PyTorch and NumPy to support reproducible execution.

% \subsection{Evaluation and Analysis}
% Each Catena module contains task-specific evaluation scripts. For segmentation, we report Variation of Information (VOI) and Adapted Rand Index Error (ARAND); for detection/classification tasks we report F1, precision, and recall. The scripts take predictions and ground truth as inputs and can aggregate results across chunks for large datasets. 

% \vspace{-0.3cm}
% \subsection{Proofreading}
% Connectomes generated using Catena modules still require manual proofreading to correct neuron trajectories and predicted synaptic connectivity. Numerous tools currently target this challenge across scales~\cite{Dorkenwald2025CAVE:Engine_edit, Zhao2018NeuTu:Reconstruction, Deng2024UniSPAC:Connectomics}. Among them, Seg2Link~\cite{Wen2023Seg2Link:Stacks} is a semi-supervised, open-source application built on \texttt{napari}~\cite{Sofroniew2025-xr} that provides manual correction utilities for resolving merge and split errors in 3D neuron segmentations. Catena integrates Seg2Link and extends it to support Zarr-backed datasets. For large-scale proofreading, Catena is adding CAVE~\cite{Dorkenwald2025CAVE:Engine_edit}- and CATMAID~\cite{Saalfeld2009CATMAID:Data}-based scripts to enable whole-brain review of segmentations, supporting merge/split operations on volumetric meshes and skeletons, respectively.

\section{CONCLUSION}
% We validated Catena's core and connectome-enricher modules on large-scale, diverse electron microscopy datasets, including out-of-distribution (OOD) volumes. The mean quantitative results on example sub-sets, summarized in Table~\ref{tab:results}, demonstrate the effectiveness of our models for foundational tasks like neuron segmentation and synapse detection across different species, as well as for detailed biological enrichment such as mitochondria segmentation and neurotransmitter classification.

Catena addresses the critical need for a comprehensive, reproducible software suite in the fragmented connectomics software landscape. Its modular architecture, consolidating core connectome extraction, connectome-enrichment and proofreading tools, supports end-to-end connectomics. Containerized runtimes, and chunk-wise processing deliver portability across environments, reproducible execution, and efficient scaling from workstations to HPC clusters. By releasing an open-source, well-documented software suite with pretrained models, Catena democratizes access to state-of-the-art methods for connectome mapping and provides a durable foundation for studies that use these mapped connectomes in downstream neuroscientific analyses.

% \textbf{Acknowledgments.} Supported by a Wellcome Trust Investigator Award to Albert Cardona (205038/Z/16/Z), MRC LMB core funding, and HHMI Janelia OSSI software-maintenance grants. We thank HHMI Janelia OSSI for integrating Catena into it; Ana Correia and Marc Corrales (MRC LMB), C.Shan Xu, and Song Pang (HHMI Janelia, Yale) for EM volume acquisition; and Daniel Han (UNSW), Michael Clayton, Nicolo Ceffa and Amina Dulac (MRC LMB), Valentin Gillet and Griffin Badalamente (Lund University) for project feedback. We are grateful to the Winding Lab (Francis Crick Institute), the Häusser Lab and Arnd Roth (UCL), and the Pape Lab (University of Göttingen) for collaborating to test Catena outside our institution.
\textbf{Acknowledgments.} Supported by a Wellcome Trust Investigator Award to Albert Cardona (205038/Z/16/Z) and the MRC LMB core funding. We thank HHMI Janelia OSSI for integrating Catena; Ana Correia and Marc Corrales (MRC LMB), C. Shan Xu and Song Pang (HHMI Janelia/Yale) for EM volume acquisition; Daniel Han (UNSW), Nicolo Ceffa and Amina Dulac (MRC LMB), Valentin Gillet, and Griffin Badalamente (Lund University) for feedback; and the Winding Lab (Francis Crick Institute), the Häusser Lab and Arnd Roth (UCL), and the Pape Lab (University of Göttingen) for external testing.

% This work is supported in part by Ministerio de Ciencia, Innovación y Universidades, AEI, under grants PID2021-126701OB-I00, MCIN/AEI/10.13039/ 501100011033, PID2019-103900GB-I00 and PID2019-10982 0RB-I00, MCIN/AEI/ 10.13039/501100011033/, cofinanced by ERDF, “A way of making Europe”, European Union - NextGenerationEU, and by grant GIU19/027 funded by the University of the Basque Country UPV/EHU. The authors declare no conflict of interest.

\textbf{Compliance with Ethical Standards.} This work is a study for which no ethical approval was required.

%\vfill
%\pagebreak

\bibliographystyle{IEEEbib} %IEEEbib
\bibliography{references_samia, references_samia_edit}

\end{document}